\documentclass[11pt]{article}
\usepackage[a4paper,margin=1in]{geometry}
\usepackage{amsmath,amssymb,amsfonts}
\usepackage{graphicx}
\usepackage{textcomp}
\usepackage{xcolor}
\usepackage{tikz}
\usepackage[ruled,vlined]{algorithm2e}
\usepackage{bbm}
\usepackage{accents}
\usepackage{contour}
\usepackage{url}
\usepackage{subcaption}
\usepackage{amsthm}
\usepackage{hyperref}

\newtheorem{remark}{Remark}

\SetKwInput{KwInput}{Input}
\SetKwInput{KwOutput}{Output}

\usetikzlibrary{positioning, arrows.meta, shapes, fit, calc, decorations.pathreplacing, intersections}

\begin{document}

\title{Strategic Federated Learning: Application to Smart Meter Data Clustering}

\author{
    M. Hassan$^{1}$, C. Zhang$^{2,3}$, S. Lasaulce$^{2,1}$, V.S. Varma$^{1}$, M. Debbah$^{2}$, M. Ghogho$^{4}$ \\
    \small{$^1$Université de Lorraine, CNRS, CRAN, F-54000 Nancy, France} \\
    \small{$^2$KU 6G Research Center, Khalifa University, Abu Dhabi UAE} \\
    \small{$^3$Central South University, Changsha, China} \\
    \small{$^4$International University of Rabat, Morocco}
}
\date{}

\maketitle

\begin{abstract}
Federated learning (FL) involves several clients that share with a fusion center (FC), the model each client has trained with its own data. Conventional FL, which can be interpreted as an estimation or distortion-based approach, ignores the final use of model information (MI) by the FC and the other clients. In this paper, we introduce a novel FL framework in which the FC uses an aggregate version of the MI to make decisions that affect the client's utility functions. Clients cannot choose the decisions and can only use the MI reported to the FC to maximize their utility. Depending on the alignment between the client and FC utilities, the client may have an individual interest in adding strategic noise to the model. This general framework is stated and specialized to the case of clustering, in which noisy cluster representative information is reported. This is applied to the problem of power consumption scheduling. In this context, utility non-alignment occurs, for instance, when the client wants to consume when the price of electricity is low, whereas the FC wants the consumption to occur when the total power is the lowest. This is illustrated with aggregated real data from Ausgrid \cite{ausgrid}. Our numerical analysis clearly shows that the client can increase his utility by adding noise to the model reported to the FC. Corresponding results and source codes can be downloaded from \cite{source-code}.
\end{abstract}

\section{Introduction}

Federated learning (FL) is an important paradigm for distributed machine learning, which has been proposed in~\cite{mcmahan-pmlr-2017}. Conventional FL involves several clients (or devices) and a central server (or fusion center -FC-). Each client uses its own data to train a local machine learning model (e.g., to learn the biases and weights of a given neural network -NN-). These local models are then sent to the FC, where they are aggregated into a global model, which is subsequently improved and shared back with the clients for further training. This process repeats, iteratively enhancing the model's performance. Compared to a centralized paradigm for which all the data would be gathered, the FL approach significantly reduces the risk of sensitive data exposure and breaches~\cite{kairouz-ftml-2021}, can implement more robust and generalizable machine learning models, can lower the costs associated with data transmission and alleviate bandwidth requirements, and can distribute the computational workload across multiple devices.

If FL is deployed from a central planner (or distributed optimization) perspective, the clients will report their model to the FC. However, in a setting that is also decentralized decision-wise, that is, when each client is left with the decision to report or not its model, it might be willing not to participate in the FL process. If the client is a wireless device, its battery level may be too low, or the network connectivity may be lost. Also, model reporting may provide a client with some model accuracy improvements that are of no effective interest to it, and may even induce a cost e.g., in terms of resources (computation, time, bandwidth, energy, money, etc.) which may be not worth it. This is one of the reasons why incentive mechanisms such as auctions have been proposed for federated learning (see e.g., \cite{cong-flpi-2020}, \cite{zhang-kde-2022}, \cite{richardson-flpi-2020}, \cite{gupta-cn-2023}).

Although synergizing FL and game theory is therefore not completely new, the present paper exploits game theory from a different perspective, namely, from a strategic information transmission standpoint (as pioneered by \cite{crawford-jeconom-1982}). In particular, this novel perspective on FL enables quantification of the amount of information a client is interested in revealing to the FC, depending on the degree of alignment between its own (final) utility function and that of the FC. The proposed framework has several new salient features to be noticed: 1. The FC produces a decision (e.g., a set of power consumption scheduling vectors for the case study investigated here) based on the model information (MI) it has from aggregating the client models; 2. The utility of each client depends on the decision taken by the FC. Each client can influence the decision only through the MI it chooses to report to the FC.; 3. Since the utility of the FC and each client can be different (non-aligned), each client adds a strategic noise to the MI that is reported. The setting of a homogeneous FL model with homogeneous IID data, where all the clients and the FC search for model accuracy maximization, corresponds to the special case of a full utility alignment scenario, in which the clients add no noise. Note that many previous works studied poisoning the model in distributed learning (e.g., \cite{fang-usenix-2020}\cite{bagdasaryan-icml-2020}\cite{bhagoji-icml-2020}), to attack the model's accuracy, additionally, there exist papers where communication noise between the clients and FC is assumed (e.g., \cite{ang-toc-2020}\cite{amiri-twc-2020}\cite{hamidi-commlett-2024}), however, the decision aspect is not considered in both cases, and the noise is not strategic, which leads to a completely different framework.

The paper is structured according to the three main contributions we provide for FL. In Sec.~\ref{sec:general-framework}, we provide the general framework of strategic FL which relies on strategic information transmission between the clients and the FC. This framework is described for unsupervised learning, namely, for the problem of data clustering. In Sec. \ref{sec:App-to-SG}, this framework is specialized further for modeling the problem of power consumption scheduling at the distribution side of electricity networks. In this scenario, clients are smart meters that report noisy cluster representatives to the FC, the latter choosing the power consumption vector for each client. In Sec. \ref{sec:Num-analysis}, we exploit real smart meter data to conduct a numerical performance analysis to illustrate the proposed framework.

\subsubsection*{\textbf{Notation}}
Throughout the paper, underlined quantities \(\underline{v}\), bold quantities \(\mathbf{M}\), calligraphic quantities \(\mathcal{X}\), \(\mathrm{Tr}()\), $\det()$ and \((\cdot)^\top\) will respectively stand for vectors, matrices, sets, trace, determinant, and the transpose operation. We will use \(\mathbb{E}\) for the expectation operator. Additionally, \(\mathcal{N}(\mu, \sigma^2)\) denotes the Gaussian distribution with mean \(\mu\) and variance \(\sigma^2\).

\section{General Strategic FL Framework Description}\label{sec:general-framework}

The conventional FL framework can be thought of as an estimation problem. Here, we add on the top of it a decision problem. We consider $M \geq 1$ clients and an FC. Each client reports its MI knowing that the FC will take a sequence of decisions based on the MI obtained by client model aggregation. This procedure is detailed next.

Each client $m\in\mathcal{M}=\{1, \dots ,M\}$ possesses a dataset of size \( N \) which is denoted by \( \mathcal{G}_m = \{\underline{\widetilde{g}}^m_1,\dots,\underline{\widetilde{g}}^m_N\} \) where \( \underline{\widetilde{g}}^m_n \in \mathcal{G}_m \subseteq \mathbb{R}^d \) represents data sample \( n \in \mathcal{N} = \{1,\ldots,N\} \), and \( d \geq 1 \) is the dimension of the convex data space \( \mathcal{G} \). Each client uses the data to train a learning model, which is represented by the (deterministic) matrix of parameters of the model $\boldsymbol{R}_m$ = $[r_m^{ij}] \in \mathbb{R}^{W \times L}$, where $W$ and $L$ are respectively the number of rows and columns. In general, this might be any learning model but, from the next section on, we will specialize our discussion and assume a data clustering model. For an NN learning model, the entries of $\boldsymbol{R}_m$ may be the weights of the NN or the gradients associated with the NN loss function. For clustering, the vectors of $\boldsymbol{R}_m$ can be the representatives of the data clusters. Also, for simplicity, we make the conventional FL homogeneity assumption which consists of assuming that all the clients implement the same model. 

Client $m$ would like the expected utility $\mathbb{E}(u_m(\underline{x}_m; \underline{g}_m))$ to be maximized, where $u_m$ is the instantaneous utility obtained when the decision taken for Client $m$ is $\underline{x}_m \in \mathbb{R}^T$, $T\geq1$, and its actual individual state is $\underline{g}_m \in \mathbb{R}^d$. The only way for Client $m$ to influence the decisions taken by the FC is to reveal in an imperfect manner the model $\boldsymbol{R}_m$ to the FC, which is modeled by a real additive noise. We assume that Client $m$ strategically reports the noisy model $\widehat{\boldsymbol{R}}_m=  \boldsymbol{R}_m+ \boldsymbol{Z}_m$ where $\boldsymbol{Z}_m$ is a realization of a noise matrix.
Here, for simplicity, it is assumed to be a Gaussian noise matrix. Denote $\underline{z}_m = \mathrm{vec}({\boldsymbol{R}_m})$, $ \mathrm{vec}$ being the vec operator. By using this notation, the probability density function of $\underline{z}_m$ writes:
\begin{align*}
\phi(\underline{z}_m; \underline{\mu}_m, \boldsymbol{\Sigma}_m) = \frac{\exp\left(\!-\frac{1}{2} (\underline{z}_m - \underline{\mu}_m)^\top \boldsymbol{\Sigma}_m^{-1} (\underline{z}_m - \underline{\mu}_m) \!\right)}{\sqrt{(2\pi)^{WL}\det(\boldsymbol{\Sigma}_m})},
\end{align*}

where $\underline{\mu}_m = \mathbb{E} (\underline{z}_m )$ and $\boldsymbol{\Sigma}_m =  \mathbb{E} (\underline{z}_m  -\underline{\mu}_m )(\underline{z}_m  -\underline{\mu}_m )^\top$. Note that other noise models can be assumed without changing the methodology proposed in this paper. For instance, a quantization noise would be more suited to model a scenario where clients only report a compressed version of their model.

For each realization of the global system state $\underline{g} = (\underline{g}_1,...,\underline{g}_M)$, the FC takes the vector of decisions $\underline{x} = (\underline{x}_1,...,\underline{x}_M)$ to maximize the following instantaneous utility:
\begin{equation}
U(\underline{x};\widehat{\underline{g}}) =
u(\underline{x}; \widehat{\underline{g}}) +
\sum_{m=1}^{M} \alpha_m u_m(\underline{x}_m;\widehat{\underline{g}}_m),
\end{equation}
where $u$ is a utility function that is common to all the clients, $\alpha_m \geq 0$ is the weight the FC applies on the individual utility $u_m$, and $\widehat{\underline{g}}_m$ is the available FC estimate of the state $\underline{g}_m$ which is obtained by using the average model 
\begin{equation}
\widehat{\boldsymbol{R}}_{\boldsymbol{Z}} = \frac{1}{M} \sum_{m=1}^M \widehat{\boldsymbol{R}}_m,
\end{equation}
where $\boldsymbol{Z} = [\boldsymbol{Z}_1 \ldots \boldsymbol{Z}_M]$. This can be written in a compact form as $\widehat{\underline{g}} = \Gamma_{\widehat{\boldsymbol{R}}_{\boldsymbol{Z}}}(\underline{g})$, $\Gamma$ being a function which may be e.g., a non-linear function (implementing an NN) or implement a clustering rule as assumed in the next sections. Model aggregation can be performed in many other ways \cite{qi-fgcs-2023}, but for simplification, in this work, we consider doing it using the Federated Averaging Algorithm (FedAVG) \cite{mcmahan-pmlr-2017}. 

The problem under consideration can be formulated as a strategic form game given by the three following components:
\begin{itemize}
\item \textbf{Players:} The set of players is $\mathcal{M} = \{1, \ldots, M\}$, corresponds to the $M$ clients;

\item \textbf{Action sets:} For Client $m$, the action is the pair $\boldsymbol{A}_m = (\underline{\mu}_m, \boldsymbol{\Sigma}_m)$ . We assume that $\forall i \in\{1,...,WL\},
\mu_m^{i} \in [\mu_{\text{min}}^{i}, \mu_{\text{max}}^{i}]$ and $\mathrm{Tr}(\boldsymbol{\Sigma}_m) \leq V_m $, $V_m\geq0$;  

\item \textbf{Utilities:} The expected utility of Client $m$, denoted by $\overline{u}_m$, encapsulates the benefit that Client $m$ derives from participating in the FL process under the chosen action $\boldsymbol{A}_m$. 
    The utility of Client $m$ is given by averaging $u_m$ over the model noises and system state realizations: 
    
    \begin{equation}  \overline{u}_m(\boldsymbol{A}_1,...,\boldsymbol{A}_m ) 
=     
\mathbb{E}_{\boldsymbol{Z},\underline{g}} \!\left[u_m\left(x_m^{\star}(\Gamma_{\widehat{\boldsymbol{R}}_{\boldsymbol{Z}}}(\underline{g}));\underline{g}_m  \right)\!\right]\!,
    \end{equation}
    where 
    \begin{equation}
        x_m^{\star}(\Gamma_{\widehat{\boldsymbol{R}}_{\boldsymbol{Z}}}(\underline{g}))
        \in \arg\max_{\underline{x}} U\left(\underline{x};
\Gamma_{\widehat{\boldsymbol{R}}_{\boldsymbol{Z}}}(\underline{g}
        )\right)\!.
    \end{equation}
\end{itemize}
\begin{remark} Note that, in general, the actions of the players do not appear explicitly in the utilities. Rather, the players' actions control the joint probability distribution over which the expectation is performed. As a consequence, the theoretical analysis of the game is non-trivial and requires a dedicated analysis which is left as an extension of this work, the main goal being here to introduce a new FL framework and illustrate its relevance for an important application. In particular, when $\overline{u}_m$ is quasi-concave with respect to $\boldsymbol{A}_m$ it is possible to prove the existence of a pure Nash equilibrium (NE) \cite{lasaulce-ap-2011}. Here, existence will only be verified ex-post namely, with simulations for the considered case study.
\end{remark}

\begin{figure}[htbp]
    \centering
    \resizebox{0.8\textwidth}{!}{
        \input{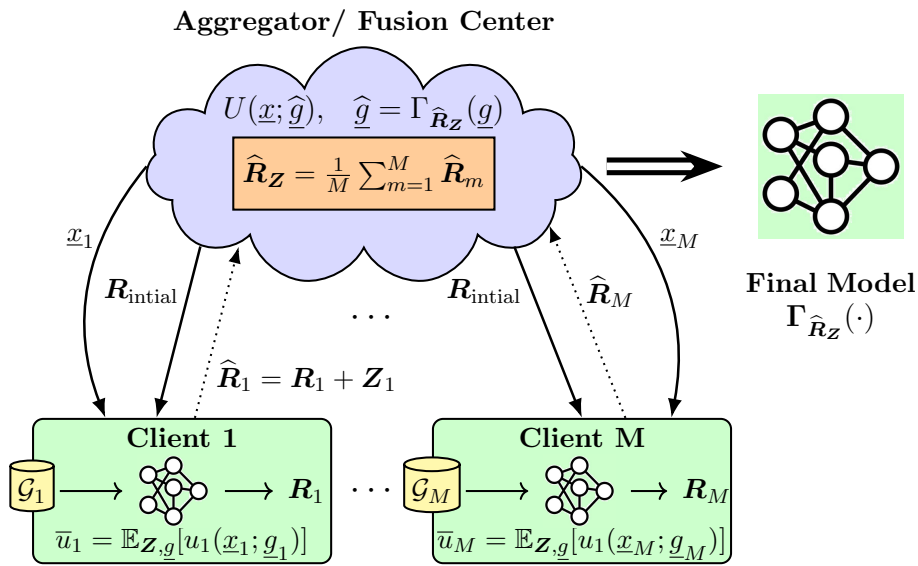}
    }
    \caption{Description of the general strategic federated learning (SFL) framework. Based on a given aggregate model information knowledge $\widehat{\boldsymbol{R}}_{\boldsymbol{Z}}$, the fusion center takes decisions $\underline{x} =(\underline{x}_1,...,\underline{x}_M)$ which maximizes its utility function $U$ and affect the clients' utility functions $\overline{u}_1,...\overline{u}_M$. The vector $\underline{g}$ represents the system state (e.g., the forecast of non-controllable consumption power or the radio channel state), and $\Gamma_{\widehat{\boldsymbol{R}}_{\boldsymbol{Z}}}(\underline{g})$ is the knowledge about the state through the learning model/function $\Gamma$. Client $m$ can influence the fusion center only through the model information reported $\widehat{\boldsymbol{R}}_m$, hence the presence of a strategic noise $\boldsymbol{Z}_m$.}
    \label{fig:system_framework}
\end{figure}

\begin{remark}
 Figure \ref{fig:system_framework} summarizes the proposed framework. Note that for the sake of clarity, the clients are assumed to send their noisy model to the FC once and for all. Therefore, in contrast with classical FL, multiple communication rounds are not assumed here. Once the FC has the model it then makes decisions.
\end{remark}

\section{Application to smart meter data clustering }\label{sec:App-to-SG}
From now on, we assume that the learning model of a client consists of an unsupervised learning model which is chosen to be a data clustering model. For Client $m$, the dataset $ \mathcal{G}_m \subseteq \mathcal{G}$, $\mathcal{G}=\mathbb{R}_{\geq 0}^d$, is partitioned into $K$ data clusters \( \mathcal{C}_1^{m} \cup \ldots \cup \mathcal{C}_K^{m} = \mathcal{G}_m \) and \( \mathcal{C}_k^{m} \cap\mathcal{C}_{k'}^{m} = \emptyset \) for any \( k \neq k' \). The representative of Cluster $\mathcal{C}_k^{m}$ is denoted by \( \underline{r}_m^k \in \mathbb{R}^d \). We assume that the model matrix $\boldsymbol{R}_m$ is precisely given by the set of representatives:
\begin{equation}
   \boldsymbol{R}_m = \left[ \underline{r}_m^1 \ldots\underline{r}_m^K \right]\!.
\end{equation}
The utility functions we choose below allow us to model several key scenarios encountered in power consumption scheduling problems within the smart grid, such as electric vehicle (EV) charging. The individual utility function for Client \( m \) is defined as:
\begin{equation}
    u_m(\underline{x}_m; \underline{g}_m) = -\|\mathbf{D}(\underline{x}_m + \delta_m \underline{g}_m)\|_{p_m},
\end{equation}
where:
\begin{itemize}
    \item \(\underline{x}_m = (x_m(1), \ldots, x_m(T))\) is assumed to be in the simplex:
    \begin{equation}
        \left\{\underline{x}_m \in \mathbb{R}^T \mid x_m(t) \geq 0,  \ \sum_t^T x_m(t) \geq E_m \right\},
    \end{equation}
    \item \(E_m \geq 0\) represents the energy need of Client \( m \),
    \item \(\mathbf{D} = \mathrm{Diag}(d(1), \ldots, d(T))\) is a diagonal matrix of size \( T \),
    \item \(\delta_m \) is either equal to \(1\) or \(-1\),
    \item \(\| \cdot \|_{p}\) denotes the \(L_{p}\)-norm given by:
    \begin{equation}
        \| \underline{v} \|_{p} = \left( |v_1|^{p} + \ldots + |v_T|^{p} \right)^{1/p}.
    \end{equation}
\end{itemize}
 In the context of power consumption scheduling (at home or for a global energy market), $\delta_m =1$, the vector $\underline{g}_m$ may represent a forecast of the non-controllable power consumption for the next $24$ hours (see e.g., \cite{beaude-ecc-2015},\cite{beaude-tsg-2016}) whereas, $\underline{x}_m$ may represent the controllable consumption vector such as the planned EV power charging vector for the home scenario. The matrix $\mathbf{D}$ may represent the prices of electricity over the $d=T$ considered time-slots. Therefore, the utility function represents the total monetary cost for consuming the energy $ \sum_{t=1}^T{x_m(t) + g_m(t)}$,  $g_m(t)\geq 0$ being the entries of $\underline{g}_m$.

The common component of the FC utility is chosen to be the following:
\begin{equation}
    u(\underline{x}; \underline{g}) = -\left\| \sum_{m=1}^{M} (\underline{x}_m+ \delta \underline{g}_m) \right\|_{p}\!\!\!,
\end{equation}
where $p\geq 1$ , $\delta \in \{-1, +1\}$. For instance, when $\delta = 1$ and $p\rightarrow \infty$ the presence of $u$ at the FC means that it wants to both minimize the network total peak consumption power (through $u$) and satisfy the consumers (through $u_m$). 

\begin{remark}
    A special instance of interest for better understanding the proposed framework is when: $M=1$, $p_m=2$, $\delta_m=1$, $p=2$, $\delta=1$, $\alpha_m=0$. That is,
\begin{equation}
    u_1(\underline{x}_1; \underline{g}_1) = -\sum_{t=1}^T D(t) (x_1(t) + g_1(t))^2,
\end{equation}
and
\begin{equation}
 U(\underline{x}; \underline{g}) =  u(\underline{x}; \underline{g})
 = - \sum_{t=1}^T (x_1(t) + g_1(t))^2,
\end{equation}
where $D(t)$ is the $t^{\text{th}}$ element of the diagonal of $\mathbf{D}$ and represents the price of electricity at time-slot $t$. In such a scenario, the client would like the total energy need $E_1$ to be waterfilled over the prices $D(1),...,D(T)$, that is, the consumption occurs mostly for low-price time-slots. On the other hand, the FC will waterfill the energy need over the non-controllable consumption $\underline{g}_1$, which means consumption will occur when the non-controllable power $g_1(t)$ is low. This perfectly illustrates the non-alignment between the decision desired by the client and the decision that maximizes the FC interest. Because of this non-alignment, the client may have an interest in only providing partial MI to the FC, which is modeled by an additive noise in this paper.    
\end{remark}

\begin{remark}
The conventional FL setting which is distortion-based can be obtained by doing the following specialization: $\forall m, \delta_m = -1$, $p_m = 2$, $\alpha_m = 0 $, $p=2$, and $\delta = -1$. In such a particular setting, the client and FC have fully aligned utilities and the clients add no noise to the model namely, $\mu_m = 0$, $v_m =0$ for all $m$. 
\end{remark}

\section{Numerical analysis}\label{sec:Num-analysis}
We test the framework
with the dataset of real measurements by Ausgrid \cite{ausgrid}. To facilitate interpretation and display the results, we first restructure the provided data (originally sampled 48 times per day), to 2 slots per day, representing average day and night consumption, and then present 48D data results. The initial clusters and representatives are obtained for each client using the classic K-means algorithm \cite{lloyd-tit-1982}. 

For the numerical results on 2D data, we use the following parameters: $p_m=1$, $\delta_m=1$, $p=\infty$, $\delta=1$, $K=2$ $\alpha_m=0$, and $E_m=1.5 $ for all $m$. With the utility functions given by
\begin{equation}
    u_m(\underline{x}_m; \underline{g}_m) = -\| \mathbf{D}_m (\underline{x}_m + \underline{g}_m)\|_1,
\end{equation}
and
\begin{equation}
 U(\underline{x}; \underline{g}) =  u(\underline{x}; \underline{g})
 = -\|  \sum_m(\underline{x}_m + \underline{g}_m)\|_{\infty},
 \label{eq:fc_utility}
\end{equation}

This scenario corresponds to the situation where the client wants to minimize the price of consumption with $\underline{x}$ denoting the flexible load (like electric-vehicle charging), while $\underline{g}$ denotes the forecast of the daily non-flexible consumption like TV, lighting, etc. as explained in the previous section. The FC, however, wants to reduce the peak consumption power. 

\subsection{2-dimensional data}
\subsubsection{Single client case}
\begin{figure}[!htbp]
    \centering
    \begin{subfigure}[t]{0.49\linewidth}
        \centering
        \resizebox{\linewidth}{!}{
        \includegraphics[width=\linewidth]{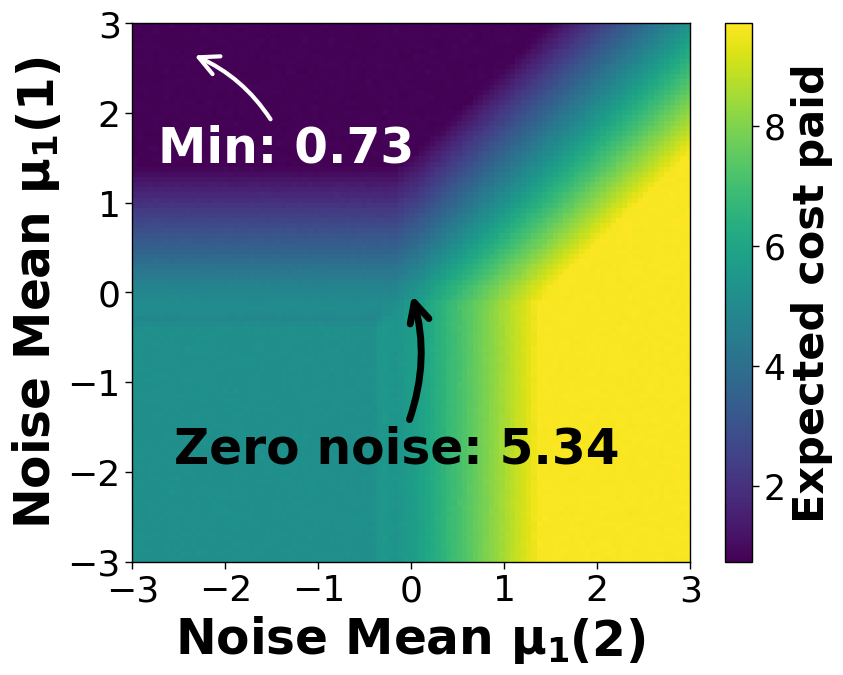}}
        \caption{Heat map of the expected cost for the client depending on the noise mean vector $\underline{\mu}_1$.}
        \label{fig:heatmap1}
    \end{subfigure}
    \hfill
    \begin{subfigure}[t]{0.49\linewidth}
        \centering
        \resizebox{\linewidth}{!}{
        \includegraphics[width=\linewidth]{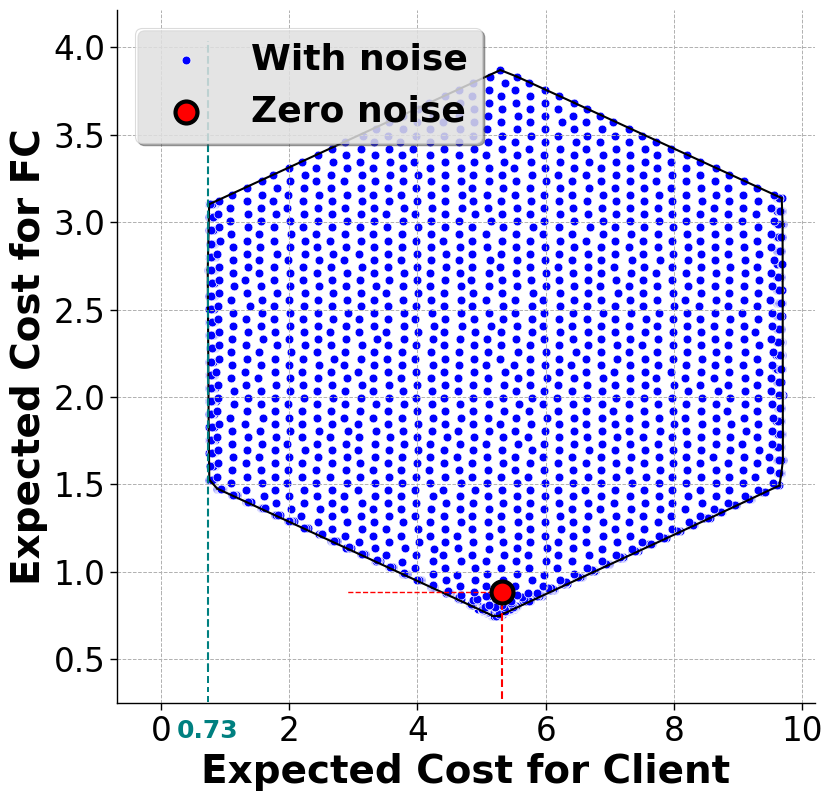}}
        \caption{Feasible cost region: client cost for abscissa axis; fusion center cost for ordinate axis.}
        \label{fig:feasible_costs_region}
    \end{subfigure}
    \caption{A single client is assumed. (a) It is seen that the client can reduce its cost from $5.34$ (when reporting its model perfectly to the fusion center, which is indicated by the black arrow) to $0.73$ by adding noise to the model reported to the fusion center, the optimal mean of the Gaussian noise being $\underline{\mu}_1=(2.650, -2.359)$. (b) The zero-noise point is Pareto dominated by the South-West orthant, which shows that the client can benefit from the noised model. Generally, most of the benefits of adding noise to the model are for the client who can make his cost as low as 0.73. However, adding a large negative noise vector to the model results in \(\Gamma_{\widehat{\boldsymbol{R}}_{\boldsymbol{Z}}}(\underline{g}) = 0\) due to clipping negative values in client's model, decreasing the fusion center's cost.}
    \label{fig:2d_1c}
\end{figure}

In this case, $M=1$, and we take $\boldsymbol{D}_1=\begin{bmatrix}
6 & 0 \\
0 & 0.1 \\
\end{bmatrix}$. The noise added by the client is $\boldsymbol{Z}_m \sim \mathcal{N}\left(\underline{\mu}_m, \mathbf{\Sigma_m}\right)
$.
Studying the results of all simulations performed, we found that a non-zero variance of the noise $\boldsymbol{Z}_m$ appears to be sub-optimal and possibly not useful. On the other hand, $\underline{\mu}_m$ plays an important role and the client is able to significantly improve his utility as seen from the heat map represented in Figure \ref{fig:heatmap1}. The heat map shows the expected cost $\mathbb{E}_{\boldsymbol{Z},\underline{g}}[-u_1(\underline{x}_1; \underline{g}_1)]
$ paid by the client w.r.t the two components of $\underline{\mu}_1$. We have also evaluated how the variance level affects performance when the mean value is fixed. Although there is a noticeable improvement in some cases, this gain is relatively small, due to space limitations, these results are not shown here. Additionally,
we illustrate in Figure \ref{fig:feasible_costs_region} the region of feasible costs when \(\mu_1(1), \mu_1(2) \in [-3, 3]^2\). This region demonstrates that the client can significantly lower their costs by introducing noise, which in turn, reduces the utility of the FC. It is important to note that the cost of the FC, as represented in the figure, is simply the opposite of its utility as defined in \eqref{eq:fc_utility}.

\subsubsection{Two-client case}
\begin{figure}[!htbp]
    \centering
    \begin{subfigure}[t]{0.49\linewidth}
        \centering
        \includegraphics[width=\linewidth]{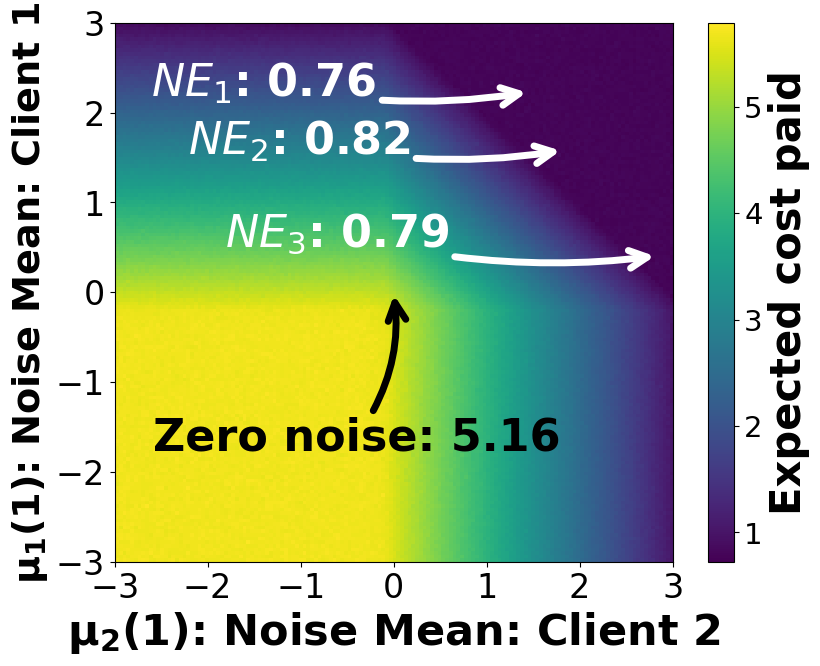}
        \caption{Expected cost of Client 1.}
        \label{fig:heatmap2}
    \end{subfigure}
    \hfill
    \begin{subfigure}[t]{0.49\linewidth}
        \centering
        \includegraphics[width=\linewidth]{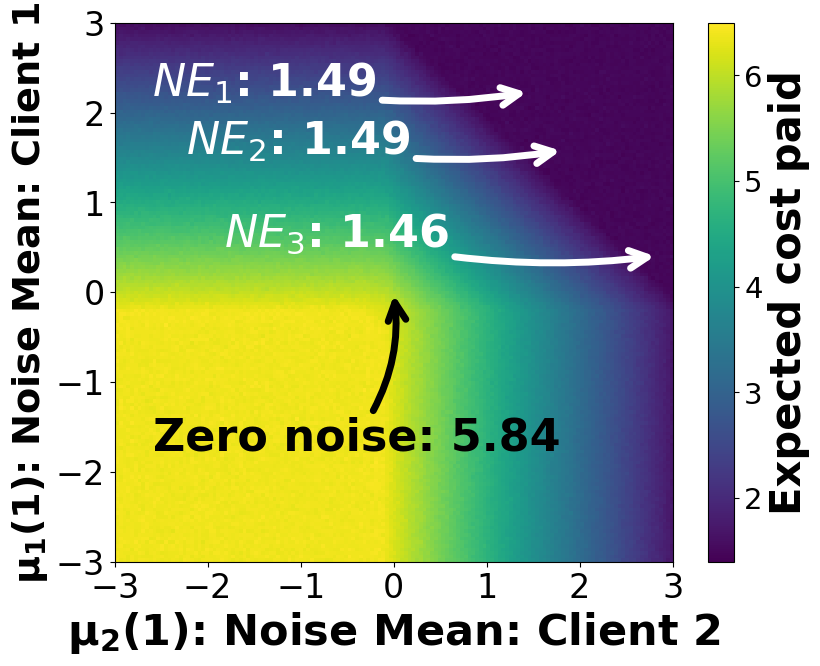}
        \caption{Expected cost of Client 2.}
        \label{fig:heatmap3}
    \end{subfigure}
    \caption{Two clients are assumed. Expected costs for Clients 1 and 2 w.r.t $\underline{\mu}_1(1)$, $\underline{\mu}_2(1)$ with 2D data. Potential Nash equilibrium (NE) points are marked at $(1.456, 2.223), (1.824, 1.581)$, and $(2.837, 0.405)$. At the NE points, the costs are much lower than at the point where the model is perfectly revealed to the fusion center.}
    \label{fig:comparison_2}
\end{figure}

The previous heatmap demonstrated that varying the first component of $\underline \mu_1$ while keeping the second fixed at 0 resulted in a cost comparable to the minimum. This motivates us to study the case of two clients by fixing $\underline{\mu}_m(2)=0$, variance $0$ and restricting the actions of each player to be $\underline{\mu}_m(1)$. We consider the same price matrix as before $\boldsymbol{D}_1=\boldsymbol{D}_2$. This configuration of the parameters allows us to plot the heatmaps of the costs associated with each client with respect to $\underline{\mu}_m(1)$, $m \in \{1,2\}$. Figures \ref{fig:heatmap2} and \ref{fig:heatmap3} illustrate this and we can exhibit Nash equilibria points for this special case as marked in the figure. As the utilities of both players are quite similar, the sum-cost at the NE is very close to that of the global minimum of the sum of the costs, i.e., we have a small price of anarchy (see e.g., \cite{lasaulce-ap-2011}).

\subsection{48-dimensional data}
\begin{figure}[!htbp]
    \centering
    \begin{subfigure}[t]{0.49\linewidth}
        \centering
        \includegraphics[width=\linewidth]{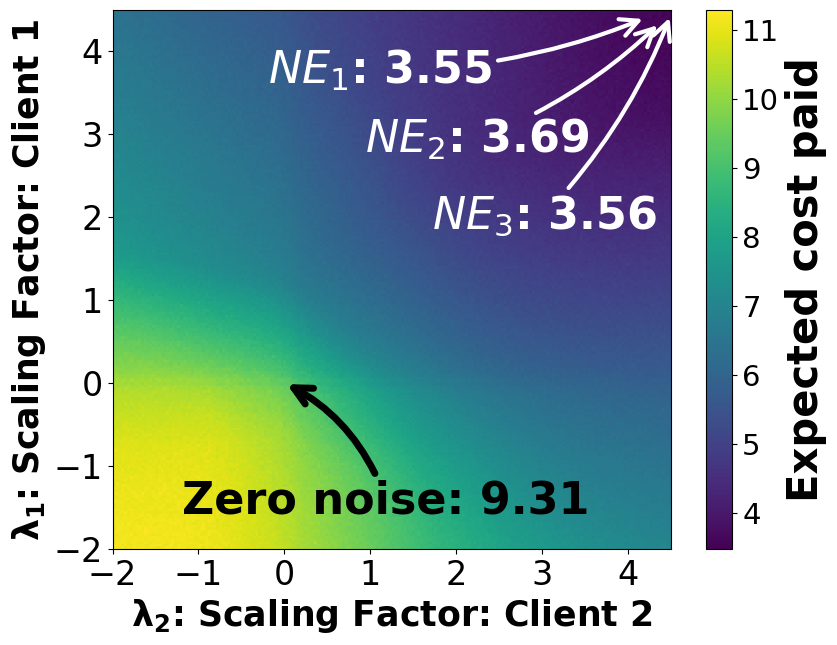}
        \caption{Expected cost of Client 1.}
        \label{fig:heatmap4}
    \end{subfigure}
    \hfill
    \begin{subfigure}[t]{0.49\linewidth}
        \centering
        \includegraphics[width=\linewidth]{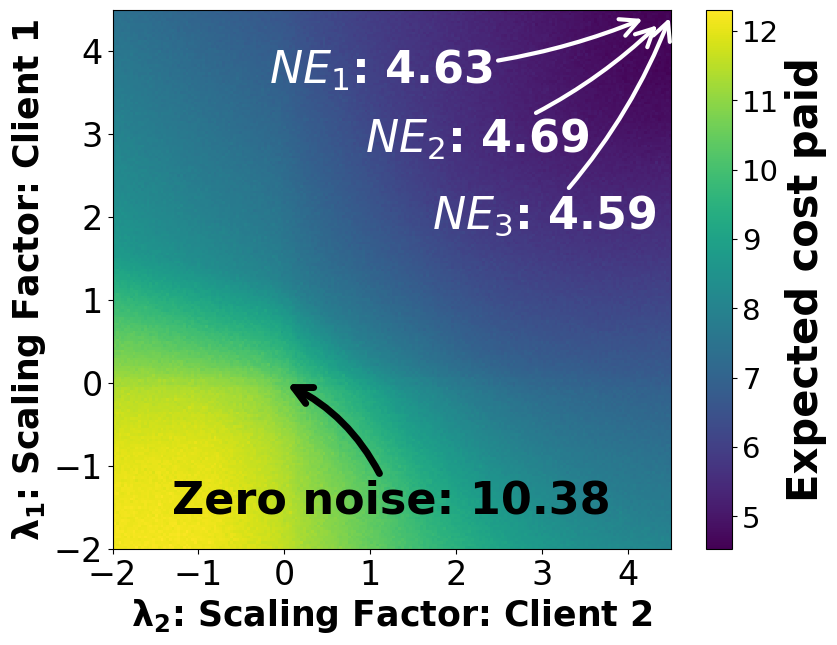}
        \caption{Expected cost of Client 2.}
        \label{fig:heatmap5}
    \end{subfigure}
    \caption{Two clients are assumed. Expected costs for Clients 1 and 2 w.r.t. $\lambda_1$, $\lambda_2$ in the 48-dimensional data case. Nash equilibrium points: $(4.204, 4.401)$, $(4.368, 4.335)$, $(4.5, 4.434)$. Costs at these points are lower than at the point with perfect model revelation to the fusion center, though the gain is less significant than in the 2D data case.}
    \label{fig:comparison_48}
\end{figure}

For the analysis on the original 48-dimensional data, we study only the two-client case. We set \( K = 4 \), \( E_m = 23 \) for all $m$, and we set \(\boldsymbol{D}_1=\boldsymbol{D}_2\) in a way to create a non-alignment between price and non-flexible consumption (price is low when consumption is high and vice versa), while keeping other parameters the same. Here, finding the optimal 48-dimensional noise vector for each client is complex and left as an extension of this work. Here we adopt a heuristic approach to check for noise vectors that decrease the expected costs paid by the clients.
For the FC, it is easy to show that maximizing \eqref{eq:fc_utility} under the energy need constraint, amounts to waterfilling over \(\underline{g}\). Consequently, in the dimensions where the values of \(\underline{g}\) are high, the corresponding \(\underline{x}\) values are low, and vice versa. 
In this scenario, clients can exploit the non-alignment between price and consumption by, for example, calculating the vector of averages along time-slots of their dataset, denoted as $\underline{g}^m_{\text{avg}}=\mathbb{E}_{\underline{g}_n\!\in\mathcal{G}_m} \left[ \underline{g}^m_n \right]$.
Clients can then set the noise vector values to be negative in the dimensions where \(\underline{g}^m_{\text{avg}}\) values are high (corresponding to low price slots) and vice versa. Specifically, each element \(\mu_m(i)\) of the noise vector \(\underline{\mu}_m\), \(i \in \{1, \ldots, 48\}\), is generated using the following heuristic:
\[
\mu_m(i) = \lambda_m \left( 0.5 - \frac{g^m_{\text{avg}}(i) - \min(\underline{g}^m_{\text{avg}})}{\max(\underline{g}^m_{\text{avg}}) - \min(\underline{g}^m_{\text{avg}})} \right)\!.
\]
Here, each element of \(\underline{g}^m_{\text{avg}}\) is normalized, shifted to include negative values, and then scaled by \(\lambda_m\) to obtain the corresponding element of \(\underline{\mu}_m \in \left[-\frac{\lambda_m}{2}, \frac{\lambda_m}{2}\right]^{48}\). Figure~\ref{fig:comparison_48} illustrates the expected costs paid by Clients 1 and 2 with respect to the scaling factors \(\lambda_m\).

\section{Conclusion}
In this paper, we have introduced a novel framework for federated learning namely, strategic federated learning. In this framework, the fusion center makes a decision based on the model information reported by the clients. In general, the utility functions of the clients and the fusion center are not aligned (i.e., the preferred decisions do not coincide). It is assumed that the only way for a client to influence the decision is through the reported model information, which leads clients to strategically add noise to the model information. The proposed framework is stated mathematically and applies to a broad class of learning models and noise models. It is applied to the important problem of data clustering for power consumption scheduling. The numerical analysis exhibits typical scenarios in which the client significantly benefits from distorting the model information appropriately. For instance, the cost associated with a given energy need can be divided by $7$ compared to the scenario in which the client perfectly reveals its model. The proposed methodology highlights the significance of strategic communication in federated learning environments, providing a detailed viewpoint on the interactions between clients and the fusion center. Even though preliminary and significant improvements have been shown with the proposed approach, there is a large space for extending our results. This includes evaluating and approximating more intricate mappings between model parameters and sophisticated tasks, more general noise models for better characterization of information disclosure, exploring higher dimensions, testing more diverse simulation scenarios, and studying different learning models.
\section*{Acknowledgments}
This work was supported by the ANR Ecorees project, the TII 6G Chair, and the Foundation Hadamard (PGMO).


\begin{thebibliography}{00}

\bibitem{ausgrid}Solar home half-hour data - 1 July 2011 to 30 June 2012, 2014,
[online] Available: \url{https://www.ausgrid.com.au/Industry/Our-Research/Data-to-share/Solar-home-electricity-data}

\bibitem{source-code} Hassan Mohamad. Strategic Federated Learning [Online]. Available: \url{https://github.com/HassanM0/SFL}

\bibitem{qi-fgcs-2023} Pian Qi, Diletta Chiaro, Antonella Guzzo, Michele Ianni, Giancarlo Fortino, and Francesco Piccialli. ``Model aggregation techniques in federated learning: A comprehensive survey.'' Future Generation Computer Systems (2023).

\bibitem{crawford-jeconom-1982}Vincent P. Crawford, and Joel Sobel. ``Strategic information transmission.'' Econometrica: Journal of the Econometric Society (1982).

\bibitem{zhang-ae-2021}Chao Zhang, Samson Lasaulce, Martin Hennebel, Lucas Saludjian, Patrick Panciatici, and H. Vincent Poor. ``Decision-making oriented clustering: Application to pricing and power consumption scheduling.'' Applied Energy 297 (2021).

\bibitem{mcmahan-pmlr-2017}Brendan McMahan, Eider Moore, Daniel Ramage, Seth Hampson, and Blaise Aguera y Arcas. ``Communication-efficient learning of deep networks from decentralized data.'' In Artificial intelligence and statistics, pp. 1273-1282. PMLR, 2017.

\bibitem{kairouz-ftml-2021}Peter Kairouz, H. Brendan McMahan, Brendan Avent, Aurélien Bellet, Mehdi Bennis, Arjun Nitin Bhagoji, Kallista Bonawitz et al. ``Advances and open problems in federated learning.'' Foundations and trends® in machine learning 14, (2021).

\bibitem{cho-arxiv-2020}Yae Jee Cho, Jianyu Wang, and Gauri Joshi. ``Client selection in federated learning: Convergence analysis and power-of-choice selection strategies.'' arXiv preprint arXiv:2010.01243 (2020).
\bibitem{lloyd-tit-1982}Stuart Lloyd. ``Least squares quantization in PCM.'' IEEE Trans. on information theory (1982).
\bibitem{cong-flpi-2020}Mingshu Cong, Han Yu, Xi Weng, and Siu Ming Yiu. ``A game-theoretic framework for incentive mechanism design in federated learning.'' Federated Learning: Privacy and Incentive (2020).
\bibitem{zhang-kde-2022}Lefeng Zhang, Tianqing Zhu, Ping Xiong, Wanlei Zhou, and S. Yu Philip. ``A robust game-theoretical federated learning framework with joint differential privacy.'' IEEE Trans. on Knowledge and Data Engineering (2022).
\bibitem{richardson-flpi-2020}Adam Richardson, Aris Filos-Ratsikas, and Boi Faltings. ``Budget-bounded incentives for federated learning.'' Federated Learning: Privacy and Incentive (2020).
\bibitem{gupta-cn-2023}Rajni Gupta, and Juhi Gupta. ``Federated learning using game strategies: State-of-the-art and future trends.'' Computer Networks 225 (2023).
\bibitem{lasaulce-ap-2011}Samson Lasaulce, and Hamidou Tembine. Game theory and learning for wireless networks: fundamentals and applications. Academic Press, 2011.
\bibitem{fang-usenix-2020}Minghong Fang, Xiaoyu Cao, Jinyuan Jia, and Neil Gong. ``Local model poisoning attacks to {Byzantine-Robust} federated learning.'' In 29th USENIX security symposium (USENIX Security 20), 2020.
\bibitem{bagdasaryan-icml-2020}Eugene Bagdasaryan, Andreas Veit, Yiqing Hua, Deborah Estrin, and Vitaly Shmatikov.``How to backdoor federated learning.'' In International conference on artificial intelligence and statistics. PMLR, 2020.
\bibitem{bhagoji-icml-2020}Arjun Nitin Bhagoji, Supriyo Chakraborty, Prateek Mittal, and Seraphin Calo. ``Analyzing federated learning through an adversarial lens.'' In International Conference on Machine Learning. PMLR, 2019.
\bibitem{ang-toc-2020}Fan Ang, Li Chen, Nan Zhao, Yunfei Chen, Weidong Wang, and F. Richard Yu. ``Robust federated learning with noisy communication.'' IEEE Trans. on Communications (2020).
\bibitem{hamidi-commlett-2024}Shayan Mohajer Hamidi, and Oussama Damen. ``Fair wireless federated learning through the identification of a common descent direction.'' IEEE Communications Letter (2024).
\bibitem{amiri-twc-2020}Mohammad Mohammadi Amiri, and Deniz Gündüz. ``Federated learning over wireless fading channels.'' IEEE Trans. Wireless Commun. (2020).
\bibitem{beaude-ecc-2015}Olivier Beaude, Samson Lasaulce, Martin Hennebel, and Jamal Daafouz. ``Minimizing the impact of EV charging on the electricity distribution network.'' In 2015 European Control Conference (ECC). IEEE, 2015.
\bibitem{beaude-tsg-2016}Olivier Beaude, Samson Lasaulce, Martin Hennebel, and Ibrahim Mohand-Kaci. ``Reducing the impact of EV charging operations on the distribution network.'' IEEE Trans. on Smart Grid (2016).

\end{thebibliography}
\end{document}